\documentclass[10pt, conference]{IEEEtran}
\usepackage{cite}
\usepackage{amsmath,amssymb,amsfonts}
\usepackage{algorithmicx,algpseudocode}
\usepackage{graphicx}
\usepackage{textcomp}
\usepackage[table]{xcolor}
\usepackage{threeparttable}
\usepackage{booktabs}
\usepackage{multirow}

\usepackage{dblfloatfix} 
\usepackage{url}
\usepackage{comment}
\usepackage{adjustbox}
\usepackage{tabularx}
\usepackage{caption} 
\newcolumntype{M}[1]{>{\centering\arraybackslash}m{#1}}
\usepackage[ruled,vlined]{algorithm2e}

\def\BibTeX{{\rm B\kern-.05em{\sc i\kern-.025em b}\kern-.08em
    T\kern-.1667em\lower.7ex\hbox{E}\kern-.125emX}}

\makeatletter
\def\ps@IEEEtitlepagestyle{
  \def\@oddfoot{\mycopyrightnotice}
  \def\@evenfoot{}
}
\def\mycopyrightnotice{
  {\footnotesize \hfill}
  \gdef\mycopyrightnotice{}
}
\makeatother

\begin{document}

\title{GraPhSyM: Graph Physical Synthesis Model
}

\makeatletter
\newcommand{\linebreakand}{%
  \end{@IEEEauthorhalign}
  \hfill\mbox{}\par
  \mbox{}\hfill\begin{@IEEEauthorhalign}
}
\makeatother
\author{
  \IEEEauthorblockN{Ahmed Agiza}
  \IEEEauthorblockA{
    \textit{Brown University}\\
    Providence, RI, USA \\
    {\tt\small ahmed\_agiza@brown.edu}}
  \and
  \IEEEauthorblockN{Rajarshi Roy}
  \IEEEauthorblockA{
    \textit{NVIDIA}\\
    Santa Clara, CA, USA \\
    {\tt\small rajarshir@nvidia.com}}
  \and
  \IEEEauthorblockN{Teodor Dumitru Ene}
  \IEEEauthorblockA{
    \textit{NVIDIA}\\
    Santa Clara, CA, USA \\
    {\tt\small tene@nvidia.com}}
  \linebreakand 
  \IEEEauthorblockN{Saad Godil}
  \IEEEauthorblockA{
    \textit{NVIDIA}\\
    Santa Clara, CA, USA \\
   {\tt\small sgodil@nvidia.com}}
      \and
  \IEEEauthorblockN{Sherief Reda}
  \IEEEauthorblockA{
    \textit{Brown University}\\
    Providence, RI, USA \\
    {\tt\small sherief\_reda@brown.edu}}
      \and
  \IEEEauthorblockN{Bryan Catanzaro}
  \IEEEauthorblockA{
    \textit{NVIDIA}\\
    Santa Clara, CA, USA \\
    {\tt\small bcatanzaro@nvidia.com}}
}


\maketitle

\begin{abstract}
In this work, we introduce GraPhSyM, a Graph Attention Network (GATv2) model for fast and accurate estimation of post-physical synthesis circuit delay and area metrics from pre-physical synthesis circuit netlists. Once trained, GraPhSyM provides accurate visibility of final design metrics to early EDA stages, such as logic synthesis, without running the slow physical synthesis flow, enabling global co-optimization across stages. Additionally, the swift and precise feedback provided by GraPhSyM is instrumental for machine-learning-based EDA optimization frameworks. Given a gate-level netlist of a circuit represented as a graph, GraPhSyM utilizes graph structure, connectivity, and electrical property features to predict the impact of physical synthesis transformations such as buffer insertion and gate sizing. When trained on a dataset of 6000 prefix adder designs synthesized at an aggressive delay target, GraPhSyM can accurately predict the post-synthesis delay (98.3\%) and area (96.1\%) metrics of unseen adders with a fast 0.22s inference time. Furthermore, we illustrate the compositionality of GraPhSyM by employing the model trained on a fixed delay target to accurately anticipate post-synthesis metrics at a variety of unseen delay targets. Lastly, we report promising generalization capabilities of the GraPhSyM model when it is evaluated on circuits different from the adders it was exclusively trained on. The results show the potential for GraPhSyM to serve as a powerful tool for advanced optimization techniques and as an oracle for EDA machine learning frameworks.
\end{abstract}

\section{Introduction}
\label{sec:introduction}

In modern Electronic Design Automation (EDA) flow, a digital design undergoes a series of transformative stages, including logic synthesis, technology mapping, physical design, and synthesis. These transformations aim to optimize the quality of results (QoR) metrics that are central to digital design, such as circuit delay, area, and power \cite{macmillen2000industrial}.  Given the potential for substantial QoR metric alterations at each stage, early-stage optimizations based solely on immediate metrics can be suboptimal, as these metrics may not accurately represent the state of the design after subsequent stages. However, since these stages are typically time and computationally-intensive, it is typically infeasible for earlier stages to optimize a design while running later stages in-the-loop for accurate feedback on QoR metrics. Thus, fast and accurate estimation of QoR metrics for later EDA stages, such as physical design and synthesis, is critical for improved optimization of a design in earlier stages, such as logic synthesis and technology mapping.

\begin{figure}[t]
\centering
\includegraphics[width=0.4999\textwidth]{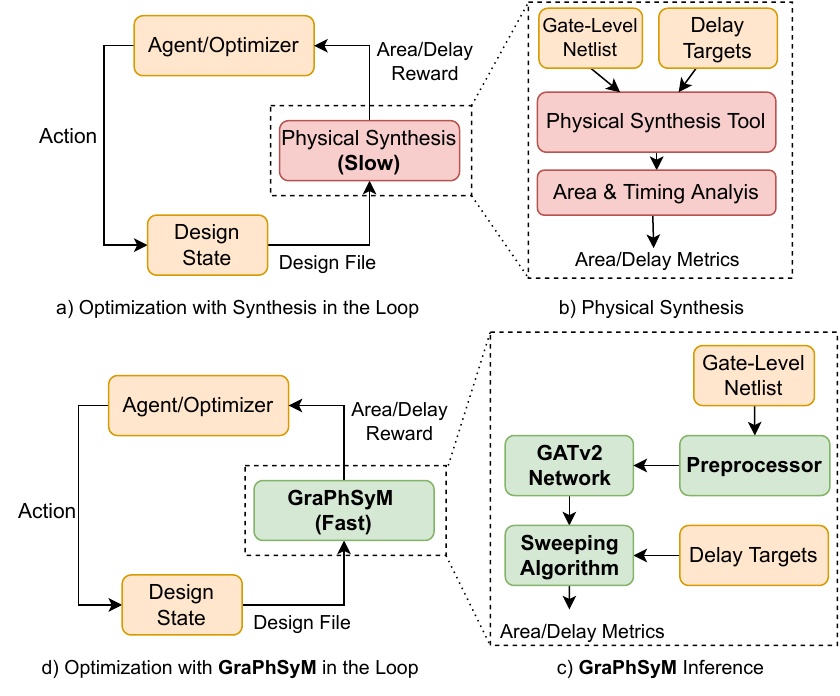}
\caption{(a) Traditional optimization/training loop. (b) Physical synthesis flow for metrics estimation. (c) GraPhSyM inference for predicting design metrics. (d) Optimization/training using GraPhSyM in the loop.}
\label{fig:overview}
\end{figure}

Several studies, such as DRiLLS \cite{hosny2020drills} and PrefixRL \cite{roy2021prefixrl}, have demonstrated that logic synthesis optimizations with final QoR metrics feedback can yield significantly better optimization than heuristic algorithms with proxy metrics feedback. DRiLLS trains reinforcement learning (RL) agents to apply logic synthesis steps in ABC \cite{brayton2010abc} in order to minimize the circuit delay and area of general logic. On the other hand, PrefixRL trains reinforcement learning (RL) agents to generate (parallel)-prefix circuits that have Pareto-optimal delay and area. Both methodologies emphasize the importance of direct QoR metric feedback, as opposed to proxy metrics such as And-Inverter Graph (AIG) \cite{biere2007aiger} or prefix node depth and count, in driving their effectiveness. PrefixRL specifically demonstrates that feedback of QoR metrics after physical synthesis optimizations with OpenPhySyn \cite{agiza2020openphysyn} in-the-loop is essential since at aggressive delay targets, a circuit that has better QoR pre-synthesis may have worse QoR post-synthesis.  However, this approach comes with a significant computational cost. For instance, physical synthesis for circuits like 32b prefix adders demands an average of 17 seconds, underscoring the need for more efficient ways to incorporate post-synthesis feedback into early-stage optimizations.

Hence, we propose GraPhSyM, a supervised learning approach where a Graph Attention Network (GATv2) \cite{velickovic2017graph,brody2021attentive} model quickly and accurately estimates post-synthesis circuit delay and area from the pre-synthesis circuit netlist (Figure \ref{fig:overview}). Our focus centers on prefix adder circuits due to their significant role in high-performance computing and machine learning ~\cite{milos_book,swartzlander_book,dadiannao,addernet}. Furthermore, a vast amount of prefix adders with a variety of logic levels and fanouts can be generated for training and evaluation. We train GraPhSyM to model the effects of OpenPhySyn's physical synthesis optimizations, such as timing-driven buffer insertion and gate sizing. However, with additional training data, GraPhSyM is designed to be extended to any circuit or synthesis tool. The key contributions of this work are:

\begin{itemize} 
\item GraPhSyM: a supervised learning approach that trains a Graph Attention Network (GATv2) model \cite{velickovic2017graph,brody2021attentive} to quickly and accurately estimate post-physical-synthesis delay and area metrics of a circuit from its pre-synthesis gate-level netlist.
\item A novel label annotation and prediction approach that associates the effect of physical synthesis transformations (area and delay change) to graph nodes (pre-synthesis netlist cell pins). This approach enables a large number of training labels (per pin) instead of just an overall area and delay labels per netlist. Furthermore, the delay change labels allow for local predictions, whereas delay labels would require the network to make predictions from the global topological structure of the netlist only, which is a more complicated task to learn.
\item An algorithm for constructing the overall area and delay metrics from the node-level predictions and a sweep algorithm for constructing overall area and delay metrics for delay targets more relaxed than the model training delay target.
\item Demonstrating the effectiveness of the GraPhSyM with a dataset of prefix adder circuits synthesized at a fixed aggressive delay target. The model inference takes 0.22s, compared to 17s for running physical synthesis, and produces accurate predictions on unseen adders (delay (98.2\%), area (97.8\%)) and on unseen delay targets that were not trained for (delay (98.3\%), area (96.1\%)). Without any additional training, the model shows promising generalization when evaluated on other circuits and produces more accurate predictions than pre-synthesis metrics.
\end{itemize}

The rest of the paper is organized as follows: Section~\ref{sec:related} discusses existing related work. Section~\ref{sec:background} explains the preliminary background and outlines our motivation. Section~\ref{sec:impl} explains our architecture and methodology. Section~\ref{sec:results} showcases the experimental evaluation of the flow, and we conclude in Section~\ref{sec:conclusions}.

\subsection{Problem Formulation}
Formally, our problem can be framed as follows: given a design's gate-level netlist DAG and early-stage EDA metrics, the objective is to train a graph neural network capable of accurately estimating the anticipated metrics for each node at later stages of the EDA flow.

\section{Related Work}
\label{sec:related}
There have been several endeavors to integrate machine-learning approaches to assist in EDA metric calculation. For instance, Dai \textit{et al.} propose a machine-learning framework for estimating an array of quality of results (QoR) metrics in high-level synthesis (HLS) flow for field-programmable gate array (FPGA) \cite{dai2018fast}. In a similar vein, Makrani \textit{et al.} presents another supervised learning model to estimate the timing and resource usage for HLS \cite{makrani2019pyramid}. However, both approaches focus on HLS flow (which is different from the traditional EDA flow) and only apply to FPGAs. In addition, they rely on minimal features from synthesis reports that might not capture essential features of the design's structure. In a different direction, the work by Kahng \textit{et al.} suggests a predictive model to reduce the divergence between two timing analysis techniques: graph-based analysis (GBA), which is a faster, less accurate analysis, and path-based analysis (PBA), which is 4x slower but provides a more accurate estimate \cite{kahng2018using}. Other research directions investigate using neural network models to provide more accurate models for cell characterization instead of look-up tables \cite{turmaristatic}\cite{raslan2023deep}, which shows the potential for neural networks to learn about cell characterization and timing analysis. Additionally, the work by Kahng \textit{et al.} shows a learning-based approach to correlate wire slew and delay from the early analysis to the results from the sign-off STA tool \cite{kahng2013learning}. A similar methodology is presented by Cheng \textit{et al.} to use machine learning models for wire timing estimation \cite{cheng2020fast}. Finally, Neto \textit{et al.} introduce an innovative approach modeling the logic paths as sentences, with the gates being a bag of words in an attempt to use ML to bridge the gap between the logic synthesis tool and the physical design \cite{neto2021read}.

\section{Motivation \& Background}
\label{sec:background}
\subsection{EDA Metrics Variance}
The Electronic Design Automation (EDA) flow represents the comprehensive process of designing, simulating, and verifying electronic systems according to the foundry's design rules \cite{wang2009electronic}. This flow is divided into several design stages, each with distinct quality metric requirements. Two principal metrics are delay, which characterizes the clock frequency at which the design can function, and cell area, which denotes the total cell area and significantly influences performance and power. An underestimation of delay can lead to timing issues, where the design may not function as intended. It could lead to scenarios where the signal does not reach the intended parts of the circuit in time, causing incorrect computations or system instability. Overestimating the delay, however, might result in overly conservative designs, thereby not fully utilizing the design's potential performance. Similarly, inaccuracies in area estimation can have implications on the cost and feasibility of the design. Underestimation could lead to insufficient allocation of resources during the design phase, causing manufacturing issues, while overestimation could unnecessarily increase production costs and result in the wastage of valuable chip real estate. These design metrics are compiled across various stages of the flow and play a crucial role in assessing the design's quality and directing the different optimization phases. As such, they are vital evaluation criteria for any EDA framework. Consequently, numerous machine-learning frameworks for EDA optimization depend on these metrics to guide or assess the optimization, a process repeated throughout the training phase. However, there exists a marked variance between the more accurate metrics (evaluated at later stages, such as physical synthesis) and the early metrics (evaluated at early stages, like logic synthesis). As a result, dependence on early metrics can lead the optimization towards suboptimal solutions since the evaluation post-physical synthesis generates different results. Thus, early metrics prove insufficient in steering optimization frameworks toward optimal results based on EDA metrics.

\begin{figure*}[b]
\centering
\includegraphics[width=0.95\textwidth]{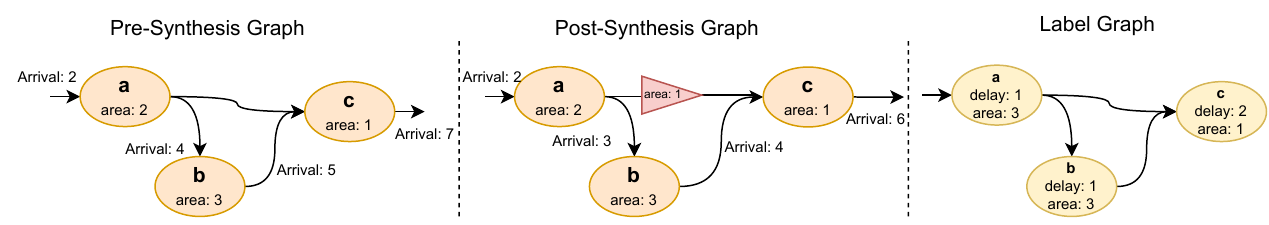}
\caption{Example of node labeling.}
\label{fig:labeling}
\end{figure*}

\subsection{Physical Synthesis Optimizations}
Physical synthesis involves a set of intricate transformations to ensure that a digital design adheres to the specified design goals. These objectives include but are not limited to timing, power, and area. Physical synthesis is critical because it translates the logical representation of the circuit into a physical layout that can be fabricated on a semiconductor wafer. Physical synthesis algorithms leverage various optimization techniques to minimize power consumption, reduce the circuit's area, and meet timing constraints. However, these goals often contradict one another, necessitating trade-off decisions during the synthesis process. Considering the complexity of physical synthesis, it is evident that efficient and accurate prediction models can greatly aid in the early design stages. Such models could provide designers with an early insight into the post-synthesis performance of the design, helping them to make informed decisions and adjustments in the preliminary stages. This is where machine learning methods, such as the one introduced in this paper, come into play. By accurately predicting post-synthesis metrics based on pre-synthesis netlists, these models can significantly streamline the EDA design process and contribute to improved design outcomes.

\subsection{Graph Attention Networks}
Machine learning offers rich opportunities to enhance and assist EDA tools and algorithms \cite{huang2021machine}. Simultaneously, the standard method for modeling EDA design is using Directed Acyclic Graphs (DAG) to represent the design's structure. Thus, machine learning frameworks that work on graphs \cite{wu2020comprehensive} are advantageous for EDA flows since they learn the complex relations among the design's graph structure. GAT \cite{velickovic2017graph} is one of the most popular types of Graph Neural Networks (GNNs). They operate on graph-structured data utilizing attention modules to overcome shortcomings of prior graph neural networks such as Graph Convolutions Networks (GCNs) \cite{kipf2016semi}. GATv2 \cite{brody2021attentive} provides further improvements over GAT by fixing limitations of static attention of standard GAT layers. We build our model upon GATv2 to capture the design features and graph structure for predicting design metrics. We also explain our techniques to enhance the graph's feature modeling and facilitate the model's convergence.

\section{GraPhSyM Architecture}
\label{sec:impl}
\subsection{Input Graph \& Feature Engineering}
\label{sec:fteng}
Our graph-based supervised learning model, GraPhSyM, offers an approach for predicting design metrics from the design features themselves. Unlike many existing methodologies that draw predictions from vectorized design features \cite{dai2018fast,kahng2018using,makrani2019pyramid}\cite{kahng2013learning}, GraPhSyM places a strong emphasis on not only understanding individual design components but also on comprehending the interconnectivity of the entire design structure. In order to capture these complex relations among various nodes and their connections, we employ graph networks. The first step in our process is to convert the design into a Directed Acyclic Graph (DAG) representation. Each pin of the design is represented as an individual node in the graph. Connections between the pins and the internal cell connections are represented as edges, building a comprehensive and interconnected graph of the design. Next, we perform a fast static timing analysis on the early-stage design to estimate the metrics before the physical synthesis process takes place. This step provides us with a basic understanding of the design's performance potential, allowing for more accurate predictions later in the process. Subsequently, we annotate each node with features extracted from the timing analysis tool and the standard cell library. The features were carefully chosen for their substantial influence on guiding the physical synthesis process. In essence, the chosen features highlight those aspects of the design that most directly affect the final output metrics. The full set of features used in our model can be found in Table~\ref{table:features}. To ensure the smooth convergence of the model, we normalize all features using z-score normalization \cite{patro2015normalization}. This process brings the values of each feature into a common range, thereby reducing the chance of disproportionate influence of certain features on the model and facilitating its learning process.

\begin{table}[]
\begin{tabular}{|p{0.3\linewidth} | p{0.6\linewidth}|}
\hline
\textbf{Feature}            & \textbf{Description \& Reasoning}                                                                                                                                                                                                       \\ \hline
\textbf{Direction}          & The direction of the pin (input or output) which directs the synthesis optimization algorithm.                                                                                                                                          \\ \hline
\textbf{Delay}              & The delay contribution of the associated pin along the worst delay path. We chose cell delay and not path-based features such as arrival time to localize the features to the node, making it easier for the model to learn.            \\ \hline
\textbf{Slew}               & The input transition time, which is a critical feature since the cell delay is usually calculated as a function of input transition time and driven capacitance.                                                                      \\ \hline
\textbf{Input Capacitance}  & The capacitance of input pins, which provides information about the driven capacitance that contributes to the cell delay model.                                                                                                        \\ \hline
\textbf{Area}               & The area of the associated cell, an important feature to be able to estimate the change in the design area.                                                                                                                             \\ \hline
\textbf{Driven Capactiance} & The capacitance driven by an output pin; while this feature can be calculated from the input capacitance, we decided to add it directly to the graph since it is an essential factor for delay calculation and optimization algorithms. \\ \hline
\textbf{Fan-out}             & The number of output connections, similar to the driven capacitance, while the fan-out can be estimated from the output edges in the graph, we also chose to provide it directly instead of letting the model learn it.                  \\ \hline
\textbf{Cell Type}          & One-hot encoding of the cell type, used to provide the model with information about the different cell types.                                                                                                                           \\ \hline
\end{tabular}
\caption{GraPhSyM node features. The first column shows the node feature; the second column describes the feature and its importance for the model.}
\label{table:features}
\end{table}

\subsection{Label Graphs}
GraPhSyM aims to accurately predict two primary Electronic Design Automation (EDA) metrics: area and delay. Central to achieving this aim is the effective generation of label graphs, which serve as the training ground for our model. A label graph is essentially a representation of design labels that correspond to the area and delay metrics post-synthesis. To create such a graph for a given design, we start by running the design through an EDA flow, where it undergoes various transformations and optimizations needed for physical realization. Following the physical synthesis, the design is subjected to a static timing analysis. This analysis evaluates the design post-synthesis to determine the path delays and calculate the total area. It is a crucial step as it provides the final metrics that will serve as our labels for the graph nodes. While it might seem straightforward to label the nodes of the graph with area and delay, the challenge lies in how to best represent these metrics so that they can be effectively used during the training phase.
\begin{figure*}[t]
\centering
\includegraphics[width=0.95\textwidth]{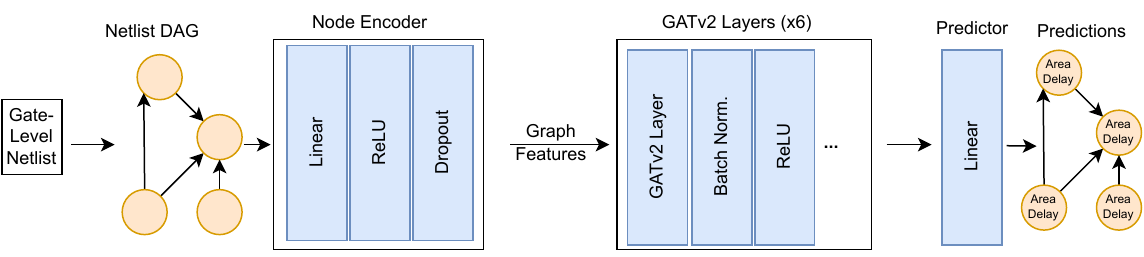}
\caption{Architecture of GraPhSyM.}
\label{fig:arch}
\end{figure*}
\begin{figure*}[t]
\centering
\includegraphics[width=\textwidth]{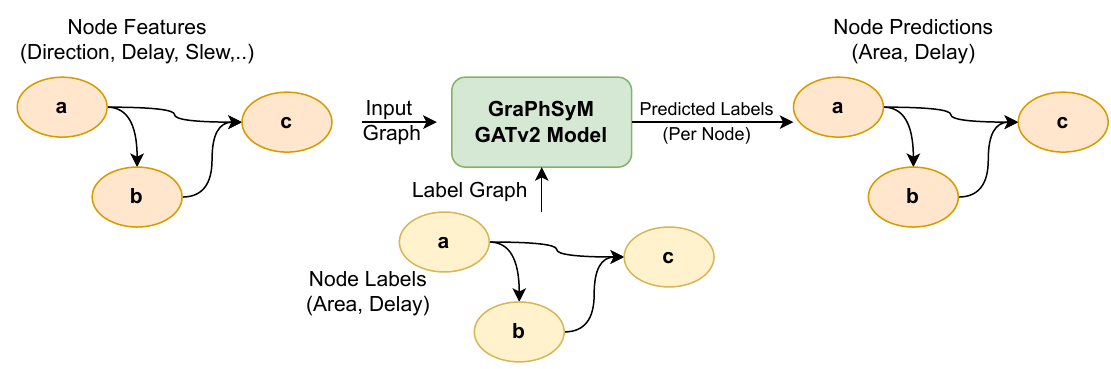}
\caption{GraPhSyM training flow.}
\label{fig:training}
\end{figure*}
\subsubsection{Delay Labeling}
There are two challenges when annotating the node delays. The first challenge pertains to the new nodes that the label graph might contain. For instance, the optimization process during physical synthesis could introduce new elements like inserted buffer trees that were absent in the original input graph. This poses a problem because if the model were to predict the new delay solely for each node present in the input graph, it would inadvertently omit the delay arising from the newly added nodes in the optimized design. Consequently, the model's ability to estimate the design's delay post-synthesis would be compromised. To counter this limitation, we deviate from the conventional method of directly using the delay feature from the analysis tool for labeling. Instead, we define the label of a node as the difference between the arrival time at the current node and the arrival time at the driving node from the original input graph. In doing so, any additional delay from new nodes added between the current node and its driver from the original graph is incorporated into the node's label. This approach ensures that the effects of any newly inserted nodes are adequately represented in the delay predictions. The second challenge involves the change in the delay for a single node before and after optimization. Oftentimes, the magnitude of this change is relatively small when compared to the delay itself. If the model attempts to predict the new delay, it is prone to essentially predict the original delay before synthesis as the new delay after synthesis, with the addition of a random delta. To illustrate, if a node has a delay of 1.2ns before synthesis and 1ns after synthesis, the model is likely to predict the delay as 1.2ns $\pm$ a small $\Delta$, given that the label is in close proximity to the feature. To address this issue, we employ a different strategy where we use the change in delay as our label instead of the delay itself. As such, in the aforementioned example, the label would be 0.2ns instead of 1ns. This enables the model to learn to predict the actual change in delay rather than falling into the trap of a one-to-one mapping of the input delay to the predicted delay.

\subsubsection{Area Labeling}
The area has a similar challenge, where the node area change is insufficient since new nodes are added. And similarly, the model needs to learn deltas to avoid the one-to-one mapping problem. Hence, for labeling the areas for each output pin in the label graph, we first add the area of all new nodes driven by the given pin. Next, we annotate the output pin with the change in the area between the pin in the input graph and its counterpart in the label graph. Figure \ref{fig:labeling} shows an example of annotating a node with delay and area labels (before being converted into deltas).

\subsection{GraPhSyM Design \& Training}

The design of GraPhSyM, as demonstrated in Figure \ref{fig:arch}, is based upon the architecture of GATv2 \cite{brody2021attentive}. At the outset, the input design is transformed into a DAG structure. Each graph node embodies a pin, while the edges signify the connections between the pins as well as the internal cell connections. We then annotate each node in the graph with a set of features derived from the timing analysis tool and the standard cell library. The next stage involves the extraction and encoding of node features. The annotated graph is fed into a node encoder layer. The role of this layer is to distill the essential attributes of each node by examining its features. This is followed by a Rectified Linear Unit (ReLU) layer, which introduces non-linearity and a dropout layer that aids in preventing overfitting. Having undergone the initial processing, the node features then traverse through six GATv2 layers. Each of these layers is equipped with eight heads and 64 hidden features. The aim of these layers is to capture and analyze the relationships between different nodes based on their connectivity, thereby leveraging the inherent graph structure of the input design. Post each GATv2 layer; the node features undergo batch normalization to standardize the features and improve model performance, ReLU activation, and another dropout layer for additional regulation. Finally, the output from the GATv2 layers is processed by a linear layer. The role of this layer is to generate two labels for each node, each indicating the changes in the node's delay and area, respectively. To obtain the final metrics, we perform an addition operation between these node labels and their corresponding features in the input graph. The delay and area across the entire design are subsequently computed by reconstructing the arrival times in the generated graph and summing the area features. At each iteration of the training process, the model tried to infer the labels for the optimized design graph. This prediction is followed by a weight update carried out by the ADAM optimizer \cite{kingma2014adam}. The optimizer employs an initial learning rate of 0.01. The process is illustrated in Figure \ref{fig:training}.

\subsection{Compositional Graph Sweeping}
The training of GraPhSyM primarily uses label graphs that have been generated through physical synthesis optimization. This training process is geared towards meeting aggressive delay targets, placing a considerable emphasis on the minimization of the design delay. If the optimization demands, this can be pursued even at the expense of a higher area cost. As a result of this strategic emphasis on high delay optimization flow during the training, the metrics inferred by GraPhSyM mirror the outcomes of an aggressive delay target.

However, this scenario does not uniformly apply to all possible applications of GraPhSyM. Specifically, when it comes to design space exploration, the requirements shift, and the framework enveloping GraPhSyM has to adjust its approach. In this scenario, the goal is to navigate the nuanced trade-off between delay and area across different delay targets. It becomes necessary to have a methodology that can accurately estimate the trade-off curve for different delay targets.

Given this, we devised an approach that makes optimal use of our feature-level predictions to estimate this trade-off. The first step involves running the design through GraPhSyM to obtain the metrics at an aggressive delay target.

These components are then fed into our sweeping algorithm described in Algorithm \ref{algo:sweep}. The algorithm is designed to use GraPhSyM predictions to emulate the physical synthesis loop, with a clear aim to estimate the delay versus area trade-off. This is achieved without the need to run the computationally-expensive physical synthesis tool, offering an efficient and effective alternative.

The algorithm commences by extracting and sorting the different delay paths present in the input graphs. We then systematically review these paths, moving from the most critical to the least critical. During this process, we swap the nodes from the input graph one by one with the corresponding nodes from the inferred graph. This swapping process continues until the path successfully meets the delay target.

This step is then repeated multiple times until all the paths have been covered or the delay targets have been met. The final output of the sweeping algorithm is a set of generated graphs. Each of these graphs represents the delay-area trade-offs for the given input graph at different delay targets. In essence, this approach allows for a comprehensive representation of the delay-area trade-offs across diverse delay targets, which can be instrumental in applications like design space exploration.

\begin{algorithm}[]
\SetAlgoLined
\KwIn{Input Graph $G1$, Inferred Graph $G2$, Target Delay $t$}
\KwResult{Optimized Graph $G1$ for the Target Delay $t$}

 Initialize $Paths$ to contain the sorted delay paths in $G1$\;

 \ForEach{path $P$ in $Paths$} {
    Set $Diff = $ delay of $P - t$\;
    \If{$Diff \leq 0$} {
        Break; \algorithmiccomment{Target already met}
    }
    
    \ForEach{output node $out_{G1}$ in $P$} {
        Set $out_{G2}$ as the equivalent node of $out_{G1}$ in $G2$\;
        Set $inp_{G1} = $ the input nodes to $out_{G1}$\;
        Set $inp_{G2} = $ the input nodes to $out_{G2}$\;
        In $G1$, replace $out_{G1}$ with $out_{G2}$\;
        In $G1$, replace $inp_{G1}$ with $inp_{G2}$\;
        Set $Diff = Diff - (\text{delay of } out_{G1} - \text{delay of } out_{G2}) - (\text{delay of } inp_{G1} - \text{delay of } inp_{G2})$\;
        \If{$Diff \leq 0$} {Break\;}
    }
}

\Return{$G1$}

 \caption{Compositional graph sweeping algorithm for predicting delay-area trade-offs at different delay targets}
 \label{algo:sweep}

\end{algorithm}

\section{Results and Discussion}
\label{sec:results}

\begin{figure*}[]
    \centering
    \begin{minipage}{\columnwidth}
        \includegraphics[width=\textwidth]{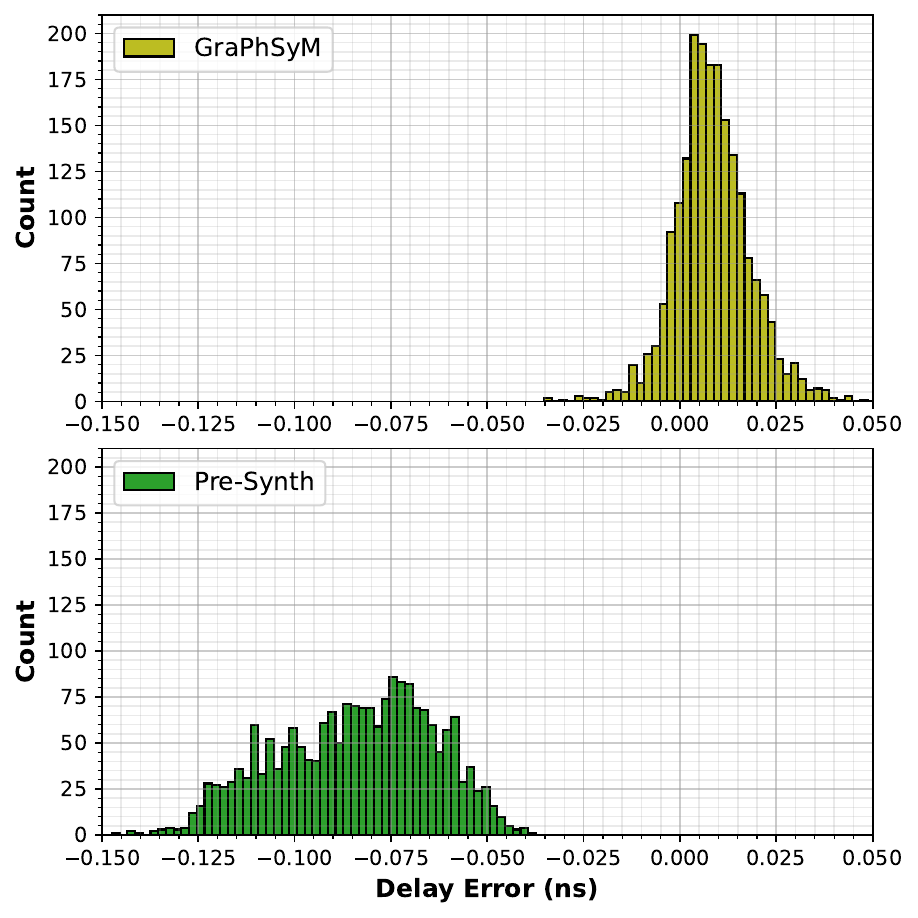}
        \caption{Distribution of differences between post-synthesis delay and GraPhSyM delay predictions (top), pre-synthesis delay (bottom).}
        \label{fig:delayhist}
    \end{minipage}\hfill
    \begin{minipage}{\columnwidth}
        \includegraphics[width=\textwidth]{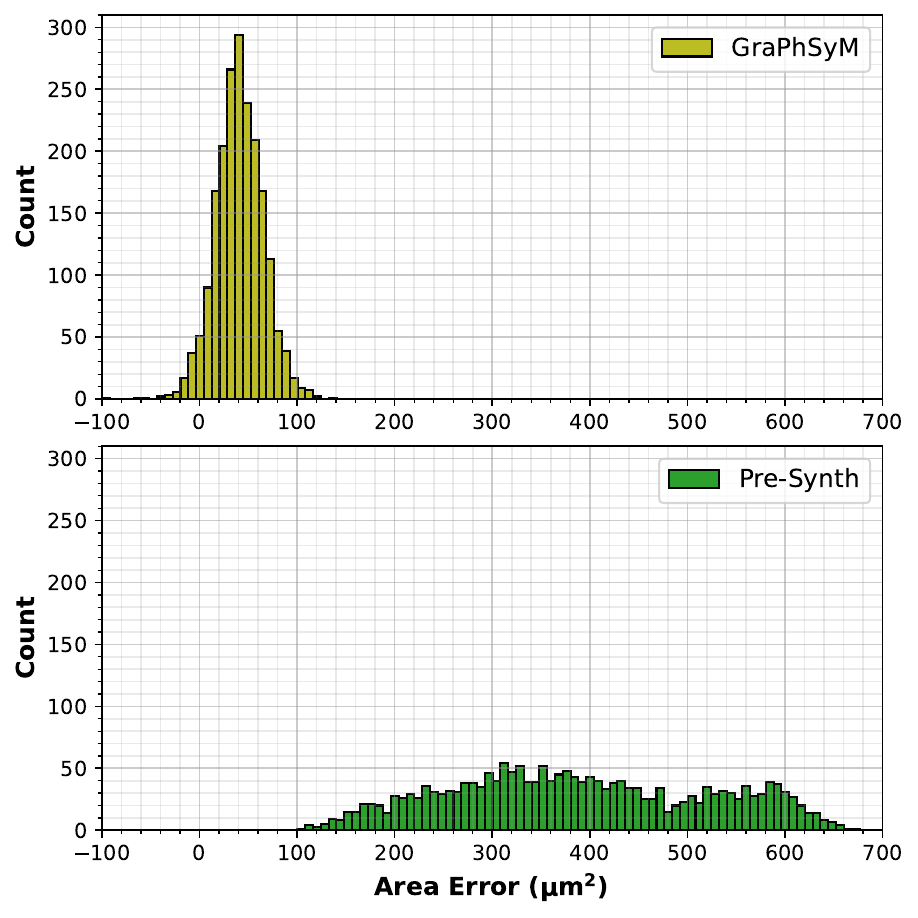}
        \caption{Distribution of differences between post-synthesis area and GraPhSyM area predictions (top), pre-synthesis area (bottom).}
        \label{fig:areahist}
    \end{minipage}\hfill
\end{figure*}

\begin{figure}[]
\centering
\includegraphics[width=0.49\textwidth]{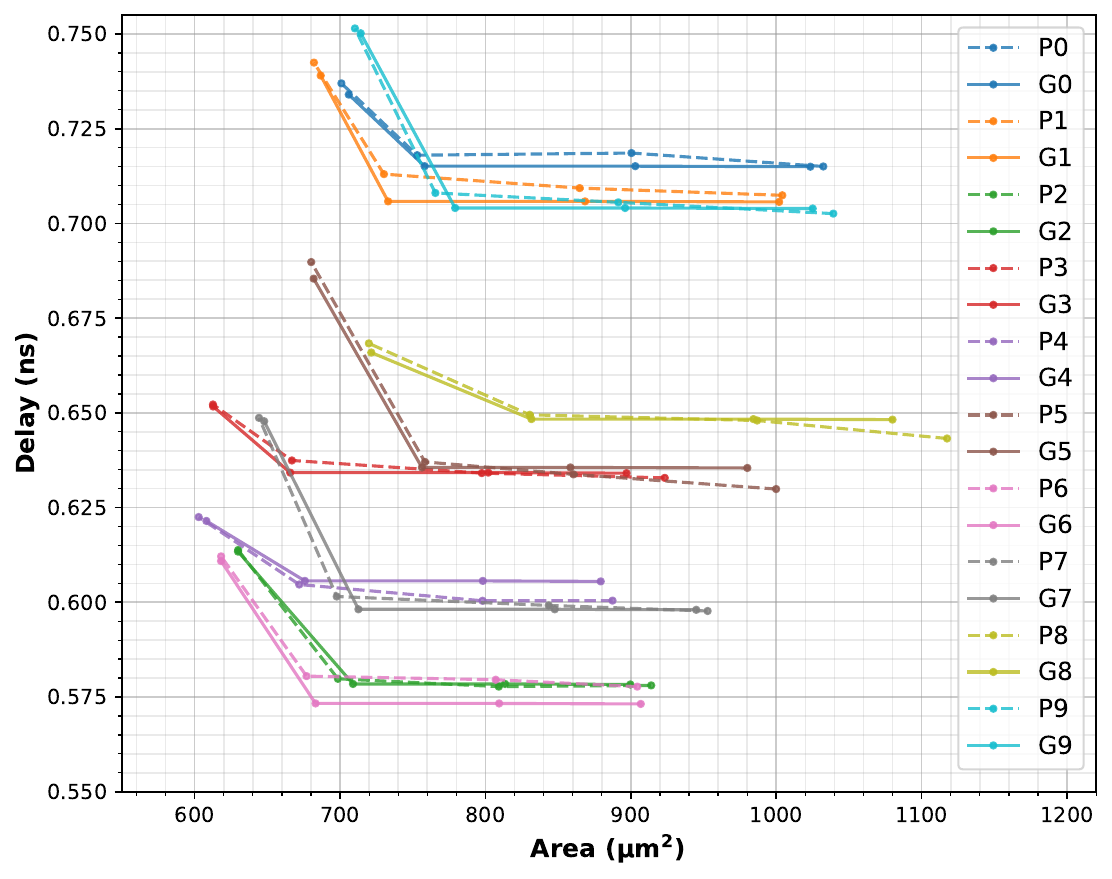}
\caption{Sweeping curves across ten adder circuits. P0-P9 are the GraPhSyM predicted values, while G0-G9 are the ground truth values for the same designs.}
\label{fig:sweep}

\end{figure}

\subsection{Dataset Preparation}
Our model is trained on a dataset comprising 10,000 instances of 32-bit prefix adders. Each adder sample is generated as follows. Initially, a prefix tree is generated with a randomized structure and a varying number of nodes. Following this step, the generated prefix tree is translated into a standard netlist and subjected to synthesis using the NanGate45 library \cite{knudsen2008nangate}. This synthesis process generates input graphs, each annotated with specific features as described in previous sections. The next stage in the dataset preparation process involves the generation of the label graph. To this end, we employ Dreamplace \cite{lin2019dreamplace} to place the generated designs. Subsequent to the placement, the designs undergo physical synthesis optimization using OpenPhySyn \cite{agiza2020openphysyn}. This optimization step serves as the precursor to the final stage, where we extract the label graphs from the optimized designs. In terms of distribution, the dataset is divided into training, validation, and testing subsets. The majority of the dataset, comprising 6,000 adders, is utilized for training the model, while 2,000 adders each are reserved for validation and testing purposes. As the prime metric for assessing the performance of our model, we use the mean absolute error (MAE), defined as: \[MAE = mean(\frac{|predicted-ground\_truth|}{ground\_truth})\]

\begin{table}[b]
\centering
\normalsize
\begin{tabular}{|l|l|l|}
\hline
\textbf{}                    & \textbf{Delay MAE} & \textbf{Area MAE} \\ \hline
\textbf{Pre-synthesis} & 12.33\%        & 34.68\%       \\ \hline
\textbf{GraPhSyM}      & 1.69\%        & 3.86\%       \\ \hline
\end{tabular}
\caption{Evaluation results of GraPhSyM on unseen adders. The top row shows the delay and area errors when using the pre-synthesis metrics to estimate the final QoR. The bottom row shows the error when using GraPhSyM to predict the final metrics post-synthesis.}
\label{table:results}
\end{table}

\subsection{Evaluation of Delay \& Area Prediction}
After training the model, we ran GraPhSyM on the evaluation dataset. GraPhSyM is significantly more accurate (Table \ref{table:results}) in predicting the post-synthesis area and delay metrics as compared to using pre-synthesis metrics. GraPhSyM predictions have an MAE of 1.69\% (delay) and 3.86\% (area). The GraPhSyM model inference takes 0.22s, compared to 17s for
running physical synthesis. Using GraPhSyM can provide almost instantaneous feedback about the post-synthesis design metrics during design space exploration instead of running a heavy and slow physical synthesis flow for each iteration. Figure \ref{fig:delayhist} shows an error distribution histogram of GraPhSyM delay predictions (top) compared to using the pre-synthesis metrics (bottom). Similarly, Figure \ref{fig:areahist} shows the error distribution histograms for area predictions.

\begin{table}[]
\centering
\normalsize
\begin{tabular}{|l|l|l|l|}
\hline
\textbf{Target}  & \textbf{Delay MAE} & \textbf{Area MAE} \\ \hline
\textbf{$T_0$}      & 1.69\%               & 3.86\%              \\ \hline
\textbf{$T_1$}      & 2.32\%               & 2.14\%              \\ \hline
\textbf{$T_2$}      & 2.97\%               & 2.41\%              \\ \hline
\textbf{$T_3$}      & 0.28\%               & 0.28\%              \\ \hline
\textbf{Average} & \textbf{1.81\%}      & \textbf{2.17\%}     \\ \hline
\end{tabular}
\caption{Evaluation of the sweeping algorithms at four delay targets for each design. $T_0$ is the minimum delay target, while $T_3$ is minimum area target.}
\label{table:sweepresults}
\end{table}

\begin{table*}[]
\normalsize
\centering
\begin{tabular}{|l|l|ll|ll|}
\hline
\multicolumn{1}{|c|}{\multirow{2}{*}{\textbf{Design}}} & \multicolumn{1}{c|}{\multirow{2}{*}{\textbf{Description}}}                   & \multicolumn{2}{c|}{\textbf{Pre-Synthesis}}                              & \multicolumn{2}{c|}{\textbf{GraPhSyM}}                                   \\ \cline{3-6} 
\multicolumn{1}{|c|}{}                                 & \multicolumn{1}{c|}{}                                                        & \multicolumn{1}{l|}{\textbf{Delay MAE}} & \textbf{Area MAE} & \multicolumn{1}{l|}{\textbf{Delay MAE}} & \textbf{Area MAE} \\ \hline
mult16                                                 & Combinational 16-bit fixed-point multiplier.                                 & \multicolumn{1}{l|}{16.99\%}              & 24.92\%             & \multicolumn{1}{l|}{13.95\%}              & 7.38\%              \\ \hline
mult32                                                 & Combinational 32-bit fixed-point multiplier.                                 & \multicolumn{1}{l|}{32.17\%}              & 31.01\%             & \multicolumn{1}{l|}{10.48\%}              & 17.77\%             \\ \hline
aes\_sbox                                              & Main module in AES \cite{daemen1999aes} encryption circuit. & \multicolumn{1}{l|}{52.30\%}              & 33.97\%             & \multicolumn{1}{l|}{5.53\%}               & 10.04\%             \\ \hline
md5\_core                                              & Main core of MD5 \cite{rivest1992rfc1321} hashing circuit.                  & \multicolumn{1}{l|}{29.84\%}              & 41.22\%             & \multicolumn{1}{l|}{3.74\%}               & 24.09\%             \\ \hline
lzd                                                    & Leading zero detector circuit.                                               & \multicolumn{1}{l|}{40.76\%}              & 20.82\%             & \multicolumn{1}{l|}{18.42\%}              & 13.28\%             \\ \hline
axmul16                                                & 16-bit approximate multiplier \cite{mahdiani2009bio}.                       & \multicolumn{1}{l|}{19.54\%}              & 21.84\%             & \multicolumn{1}{l|}{8.65\%}               & 13.06\%             \\ \hline
csela32                                                & 32-bit carry select adder.                                                   & \multicolumn{1}{l|}{27.78\%}              & 37.55\%             & \multicolumn{1}{l|}{10.14\%}              & 15.37\%             \\ \hline
priority\_enc                                          & Priority encoder binary compression circuit.                                 & \multicolumn{1}{l|}{58.92\%}              & 30.42\%             & \multicolumn{1}{l|}{6.99\%}               & 23.95\%             \\ \hline

\end{tabular}
\caption{Evaluation of GraPhSyM on eight unseen designs from different categories. The first two columns describe the design circuit. The next two columns show the error when using the pre-synthesis for metrics estimation, compared to the error when using GraPhSyM in the last two columns.}
\label{table:gen}
\end{table*}

\subsection{Evaluation of Compositional Graph Sweeping}
Next, we assess the compositional graph sweeping algorithm using the evaluation dataset. In this phase, each design in the evaluation set was subjected to physical synthesis at four distinct delay targets. These targets spanned between the maximum and minimum delay attributable to the specific design. Correspondingly, the GraPhSyM sweeping algorithm was also deployed at the same delay targets. Table \ref{table:sweepresults} captures the outcomes of the sweeping algorithm in terms of delay and area error at each of the four delay targets. The results show the delay and area error of GraPhSyM sweeping algorithm with an average Mean Absolute Error (MAE) of 1.81\% and 2.17\% for the delay and area predictions, respectively. Figure \ref{fig:sweep} provides a more visual assessment of the algorithm's performance. It presents a comparative analysis of the delay-area trade-off curves for a selection of prefix adders in the dataset alongside the performance of the GraPhSyM sweeping algorithm. These graphical representations highlight the algorithm's ability to closely follow the actual trade-off trends and deliver precise predictions for post-synthesis design metrics.

\subsection{Generalization to Unseen Design Spaces}
In order to explore GraPhSyM's versatility and adaptability across diverse design spaces, we evaluated the same model on unseen design spaces. The evaluation process involved an array of eight distinct design types that were unexplored by the model. A significant point of note is that no fine-tuning was employed during this phase, allowing us to assess the inherent capabilities of the model in new design spaces. The results, summarized in Table~\ref{table:gen}, show the prediction accuracy for delay and area in the new designs. The average MAE stands at 9.74\% for delay and 15.61\% for area predictions. In comparison, the use of pre-synthesis metrics resulted in significantly higher errors, with MAEs of 34.79\% and 30.22\% for delay and area, respectively. These findings accentuate GraPhSyM's generalizability. Despite being trained exclusively on prefix adder circuits, the model displayed the ability to infer on general circuits with varying characteristics and structures and that it has the potential to operate effectively across general circuits with a potential for higher accuracy when trained on a diverse set of circuit designs. In future work, we plan to refine our model through fine-tuning across a broader range of design spaces to further enhance the generalization capabilities for unseen design spaces.

\section{Conclusion}
\label{sec:conclusions}
In this paper, we have presented GraPhSyM, a graph-based, supervised learning approach designed for estimating digital circuit delay and area metrics post-physical synthesis phase. GraPhSyM's GATv2 neural network architecture enables it to efficiently model and quantify the impacts of various optimizations that are typically applied during the process of physical synthesis. The performance of GraPhSyM has been validated on a dataset comprising prefix adder circuits, with each circuit optimized under a fixed delay target. GraPhSyM proved adept at predicting the metrics of unseen adder circuits with significant accuracy. These predictions are fundamentally facilitated by the compositional nature of our model, which affords it the ability to make predictions at the granular level of individual nodes. This feature-level prediction capability has been pivotal in the development of an algorithm that can leverage GraPhSyM's predictions to anticipate metrics across a wider range of relaxed delay targets. An additional highlight of GraPhSyM’s performance is its demonstrated potential for generalization across a broad spectrum of circuits from unseen design spaces. The results we have presented demonstrate the potential of graph neural networks in learning the optimization patterns of physical synthesis and providing visibility of late-stage design metrics to the early stages of EDA, such as logic synthesis. GraPhSyM provides significant strides in improving the overall efficiency and effectiveness of the design process. Moving forward, our goal is to augment the capabilities of GraPhSyM further, enhancing its accuracy on general circuits. We plan to accomplish this by diversifying our training datasets, integrating GraPhSyM with logic optimization algorithms, and broadening its scope to encompass other crucial metrics, such as power. We are also keen to extend its application to other synthesis tools, thereby expanding its range of impact and utility.

{
\bibliographystyle{IEEEtran}
\bibliography{references}
}

\end{document}